\documentclass[sigconf, nonacm]{acmart}

\newcommand\vldbdoi{10.14778/3796195.3796196}
\newcommand\vldbpages{794 - 807}
\newcommand\vldbvolume{19}
\newcommand\vldbissue{5}
\newcommand\vldbyear{2026}
\newcommand\vldbauthors{\authors}
\newcommand\vldbtitle{\shorttitle} 
\newcommand\vldbavailabilityurl{https://github.com/secretflow/ACoLab/tree/main/Autodp-paper-code}
\newcommand\vldbpagestyle{empty} 

\usepackage{xeCJK}
\usepackage[utf8]{inputenc}
\usepackage{multirow}
\definecolor{shadecolor}{rgb}{0.92,0.92,0.92}
\usepackage{framed}
\usepackage{array}

\begin{document}

\title{LLM-AutoDP: Automatic Data Processing via LLM Agents for Model Fine-tuning}

\author{Wei Huang*}
\affiliation{%
  \institution{Ant Group}
  \city{Beijing}
  \state{China}
}
\email{hw378176@antgroup.com}

\author{Anda Cheng*}
\affiliation{%
  \institution{Ant Group}
  \city{Beijing}
  \state{China}
}
\email{andacheng.cad@gmail.com}

\author{Yinggui Wang$\dagger$}
\affiliation{
  \institution{Ant Group}
  \city{Beijing}
  \state{China}
}
\email{wyinggui@gmail.com}

\author{Lei Wang}
\affiliation{%
  \institution{Ant Group}
  \city{Beijing}
  \state{China}
}
\email{shensi.wl@antgroup.com}

\author{Tao Wei}
\affiliation{%
  \institution{Ant Group}
  \city{Beijing}
  \state{China}
}
\email{lenx.wei@antgroup.com}






\begin{abstract}

Large Language Models (LLMs) can be fine-tuned on domain-specific data to enhance their performance in specialized fields. However, such data often contains numerous low-quality samples, necessitating effective data processing (DP). In practice, DP strategies are typically developed through iterative manual analysis and trial-and-error adjustment. These processes inevitably incur high labor costs and may lead to privacy issues in high-privacy domains like healthcare due to direct human access to sensitive data. Thus, achieving automated data processing without exposing the raw data has become a critical challenge. 
To address this challenge, we propose \textbf{LLM-AutoDP}, a novel framework that leverages LLMs as  agents to automatically generate and optimize data processing strategies. Starting from an initial prompt, our method generates multiple candidate strategies and iteratively refines them using feedback signals and comparative evaluations. This iterative in-context learning mechanism enables the agent to converge toward high-quality processing pipelines without requiring direct human intervention or access to the underlying data.
To further accelerate strategy search, we introduce three key techniques:  
(1) \textit{Distribution Preserving Sampling}, which reduces data volume while maintaining distributional integrity;  
(2) \textit{Processing Target Selection}, which uses a binary classifier to identify low-quality samples for focused processing; and  
(3) \textit{Cache-and-Reuse Mechanism}, which minimizes redundant computations by reusing prior processing results.
We evaluate LLM-AutoDP on five medical datasets across three model architectures. Results show that models trained on data processed by our framework achieve over 80\% win rates against models trained on unprocessed data. Compared to AutoML baselines based on LLM agents, LLM-AutoDP achieves approximately a 65\% win rate. Moreover, our acceleration techniques reduce the total searching time by up to $10\times$, demonstrating both effectiveness and efficiency.

\end{abstract}

\maketitle

\pagestyle{\vldbpagestyle}

\begingroup\small\noindent\raggedright\textbf{PVLDB Reference Format:}\\
\vldbauthors. \vldbtitle. PVLDB, \vldbvolume(\vldbissue): \vldbpages, \vldbyear.\\
\href{https://doi.org/\vldbdoi}{doi:\vldbdoi}
\endgroup
\begingroup
\renewcommand\thefootnote{}\footnote{\noindent
* Equal contribution. $\dagger$ Corresponding author.  \\
This work is licensed under the Creative Commons BY-NC-ND 4.0 International License. Visit \url{https://creativecommons.org/licenses/by-nc-nd/4.0/} to view a copy of this license. For any use beyond those covered by this license, obtain permission by emailing \href{mailto:info@vldb.org}{info@vldb.org}. Copyright is held by the owner/author(s). Publication rights licensed to the VLDB Endowment. \\
\raggedright Proceedings of the VLDB Endowment, Vol. \vldbvolume, No. \vldbissue\ %
ISSN 2150-8097. \\
\href{https://doi.org/\vldbdoi}{doi:\vldbdoi} \\
}\addtocounter{footnote}{-1}\endgroup

\ifdefempty{\vldbavailabilityurl}{}{
\vspace{.3cm}
\begingroup\small\noindent\raggedright\textbf{PVLDB Artifact Availability:}\\
The source code, data, and/or other artifacts have been made available at \url{\vldbavailabilityurl}.
\endgroup
}

\section{Introduction}
The remarkable success of large language models (LLMs) has led to their widespread adoption across diverse domains~\cite{grattafiori2024llama,yang2025qwen3}. 
While applying LLMs in specialized fields such as healthcare~\cite{liu2024survey,zheng2025large}, domain-specific fine-tuning is often necessary to achieve optimal performance. 
A critical challenge in domain-specific fine-tuning lies in data acquisition: domain-specific datasets are typically collected through web scraping, crowdsourcing, or other automated methods~\cite{chen2023huatuogpt,zeng2020meddialog}, which frequently introduce substantial noise and render the raw data unsuitable for direct use. 
Consequently, data processing (DP) emerges as a pivotal step in the fine-tuning pipeline.

Traditionally, DP strategies are manually designed by iterative human observation and trial-and-error adjustments. 
While this can yield usable datasets, it suffers from two issues: (1) it incurs significant labor costs, and (2) more critically, it could lead to privacy issues, especially when handling sensitive data (e.g., medical records). 
As a result, an ideal solution is to generate DP strategies automatically. Automatically generating DP strategies mainly faces two issues: (1) Which DP operators need to be used? (2) What is the execution order of these operators?
Existing approaches towards automatic data processing typically leverage AutoML technologies~\cite{chi2024sela,feurer2022auto,shende2022cleants,Olson2016TPOTAT,Karras2023AutoMLWB,Minh2018AutomatedID}, employing optimization algorithms such as three search, Bayesian optimization, or evolutionary algorithms to identify the most effective combination of basic data processing operations. 
Despite their sophistication, these methods suffer from two major challenges: 
(1) These optimization algorithms do not comprehend the intrinsic meanings or semantics behind various DP strategies, 
which generally results in slow convergence; 
(2) They are not tailored for data processing in LLM fine-tuning tasks, thereby ignoring the problem of the large computational overhead associated with LLM data processing and fine-tuning.

To address these challenges, we propose \textbf{LLM-AutoDP} to leverage LLMs as agents for automatic data processing.
LLM-AutoDP operates iteratively through two interactive modules: {strategy generation} and {strategy evaluation}. 
The strategy generation module employs LLMs as agents to create semantically meaningful and effective DP strategies by using the initial prompt to inject domain-specific knowledge into the generation process.
The strategies are then passed to the strategy evaluation module to get feedback for further strategy refinement. 
To effectively utilize feedback signals to improve subsequent generations, we propose a group relative comparison mechanism. Specifically, we refine the prompt by combinations of multiple generated strategies and their corresponding feedback signals as in-context information, guiding the LLMs toward more effective strategy formulation.
The evaluation module applies generated strategies to the raw data, followed by model fine-tuning and testing. 
To ensure both speed and reliability of evaluation, we introduce three key techniques: 
(1) {Distribution-preserving Sampling}, to reduce the amount of data in fine-tuning while maintaining the original distribution characteristics; (2) {Processing Target Selection}, to identify the most informative subsets of data for processing; and (3) {Cache-and-Reuse Mechanism}, to minimize redundant computation by reusing previously processed results. 
Together, these components enable LLM-AutoDP to efficiently explore the space of DP strategies and converge to high-quality solutions that enhance model fine-tuning performance with maintaining computational efficiency.

We employed two state-of-the-art (SOTA) LLMs as agents and conducted extensive experiments on five medical datasets and three models. The experimental results show that, compared to models
trained on unprocessed data, the models trained on data processed by LLM-AutoDP achieve a win rate of over 80\%. Additionally, our method achieves approximately a 65\% win rate when
compared to AutoML baseline approaches based on LLM agents. Furthermore, we demonstrated through ablation experiments that our framework maintains result stability even when faced with varying numbers of strategies generated in the initial rounds, and different models used by the agents, which also proves the robustness of our method. The three acceleration strategies that we propose reduce the total
processing data time by up to $10\times$, demonstrating both effectiveness and
efficiency.

\section{Related Work}
\textbf{LLM Agent.}
The era of intelligent agents has arrived, driven by revolutionary advancements in LLMs. LLM agents, characterized by their goal-driven behaviors and dynamic adaptation capabilities, potentially represent a critical pathway toward achieving artificial general intelligence (AGI). Both academia and industry have applied LLM agents to a wide range of fields, including code generation~\cite{hong2023metagpt}, data security~\cite{li2025commercial}, social sciences~\cite{li2023econagent,ma2024understanding}, game generation~\cite{chen2023gamegpt}, tool creation~\cite{zhang2024agentcf}, model training~\cite{chi2024sela}, and more. Method MetaGPT~\cite{hong2023metagpt} encodes Standard Operating Procedures (SOPs) into prompt sequences to enable a more streamlined workflow, allowing agents with human-like domain expertise to validate intermediate results and reduce errors. SELA~\cite{chi2024sela} introduces Tree-Search Enhanced LLM Agents, an agent-based system that leverages Monte Carlo Tree Search (MCTS) to optimize the AutoML process. By representing pipeline configurations as trees, agents can conduct experiments intelligently and iteratively refine their strategies, facilitating a more effective exploration of the machine learning solution space. GameGPT~\cite{chen2023gamegpt} leverages a dual-agent collaboration and a hierarchical approach, using multiple internal dictionaries to automate and enhance the game development process. Currently, there is a notable lack of research on using LLM agents for systematic, automated processing of data for LLM fine-tuning. Our work aims to fill this gap.

\textbf{Automated Data Preprocessing with AutoML.}
Data preprocessing can be automated to varying degrees, depending on the nature of the  operations and the implementation of the underlying models. The most basic level of automation still relies on traditional hard-coded techniques~\cite{Mumuni2024AutomatedDP}, while more advanced approaches leverage AutoML technologies~\cite{BeheraAutomationID,Liu2021AutomaticDA,Dissanayake2025ASO}. Although many preprocessing methods are specifically designed to handle a single preprocessing task ~\cite{zhang2023data,jarrett2022hyperimpute}, numerous studies ~\cite{bilal2022auto,liu2021autodc,li2023diffprep,Jin2023AutoKerasAA} aim to achieve multiple preprocessing functions simultaneously. AutoPrep~\cite{bilal2022auto} performs automatic missing data imputation, data type detection, duplicate removal, categorical data encoding, feature scaling. 
Other methods AutoData~\cite{liu2021automatic}, DataAssist~\cite{goyle2024dataassist}, BioAutoMATED~\cite{valeri2023bioautomated}, Atlantic~\cite{santos2023atlantic}, and DiffPrep, are designed as dedicated preprocessing plugins that can be integrated into standard AutoML methods~\cite{He2019AutoMLAS,Shi2020SAFESA}. For example, AutoData ~\cite{liu2021automatic} implements a specialized data processing module that operates as an end-to-end mechanism within a standard AutoML framework, enabling the execution of various data preprocessing and augmentation tasks, including data acquisition, labeling, cleaning, and enhancement. These methods are typically applied to traditional models and small datasets, where dozens of iterations are often required to derive high-quality data processing strategies. However, when applied to LLMs, these methods face significant challenges.
Although there are some recent automated data processing frameworks for LLM training~\cite{dj,dj2}, these methods do not fully use LLM as an agent for fully automated strategy selection.
LLMs often require substantial time to process data, making the application of these preprocessing techniques for automatic data processing on LLMs highly time-consuming and thus impractical.

\section{Framework and Key Mechanisms}

\begin{figure*}
    \centering
    \includegraphics[width=0.77\linewidth]{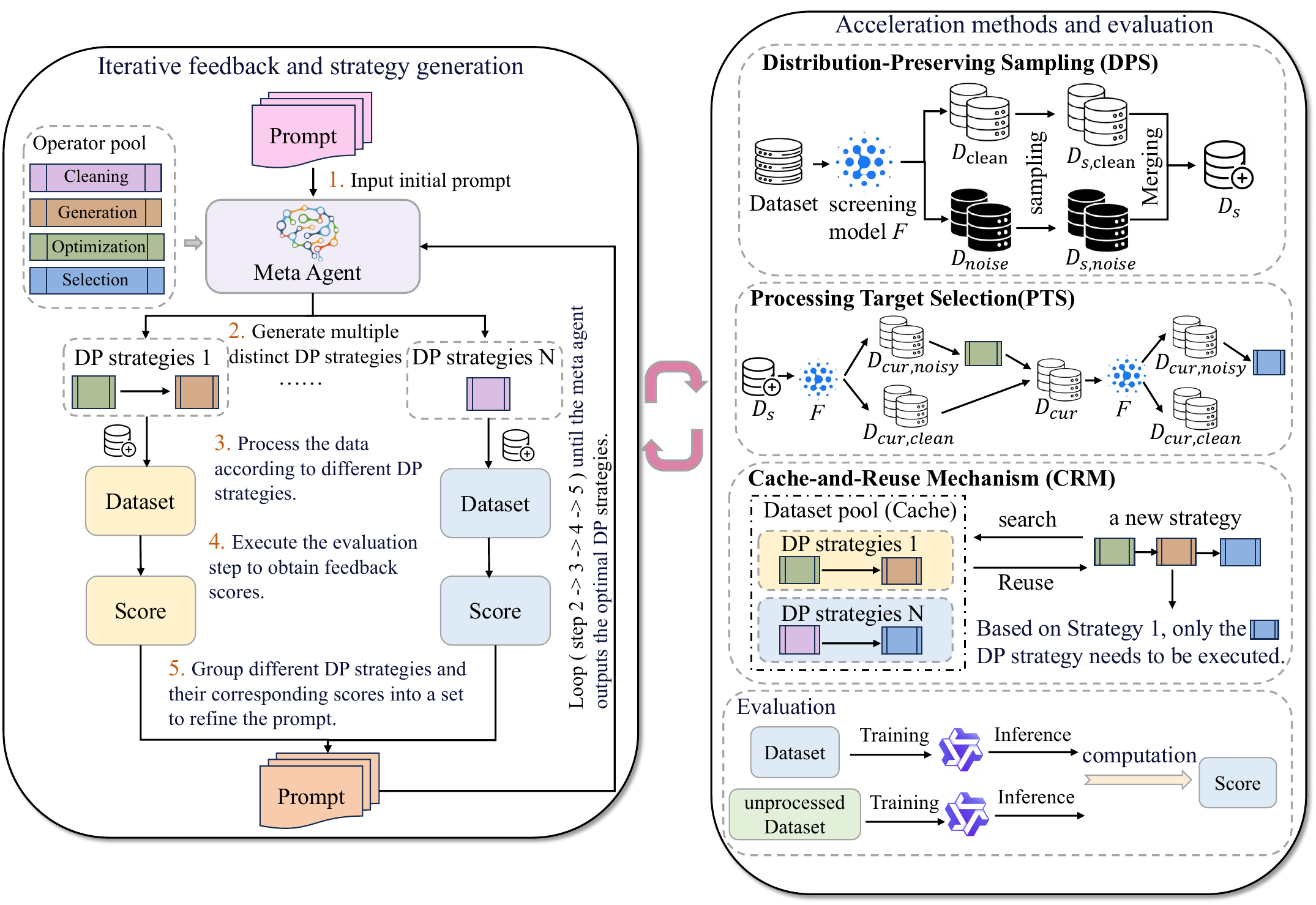}
    \caption{The overall framework of LLM-AutoDP. The left part utilizes an LLM as an agent to iteratively refine the prompt, enabling the agent to generate high-quality data processing strategies. The right part consists of an evaluation module for generating strategy feedback scores and an acceleration module for speeding up data processing. The LLM iteratively controls the interaction between the two parts to select high-quality processing strategies.}
    \label{fig1}
\end{figure*}

\subsection{Framework of AutoDP}
The overall framework of LLM-AutoDP is depicted in Figure~\ref{fig1}. 
This framework comprises two interrelated modules: (1) a strategy generation module that employs LLMs as agents to automatically design DP strategies, and (2) a strategy evaluation module that assesses the strategies through model training and evaluation. 

\begin{table}[!t]
\caption{Operator taxonomy with four categories}
\label{tab:ops}
\centering
\scalebox{0.9}
{\begin{tabular}{c|p{6cm}}
\hline
\multicolumn{1}{c|}{\textbf{Data Processing}} & \textbf{Operations} \\ \hline
\multirow{5}{*}{Cleaning} & Apply MinHashLSH to eliminate duplicate samples.~\cite{tirumala2023d4} \\
 & Remove low-level noise, such as HTML tags. \\
 & Maintain the ratio of special characters in the sample within a specific range. \\
 & Keep the number of tokens in the sample within a specific range. \\
 & Keep the sample of word-level n-gram repetition ratios within a specific range. \\ \hline
\multirow{3}{*}{Optimization} & Optimize the questions in the sample.~\cite{chen2024data}\\
 & Optimize the answer in the sample.~\cite{chen2024data} \\
 & Optimize both the questions and answers in the sample.~\cite{chen2024data} \\ \hline
\multirow{3}{*}{Generation} & Generate the missing questions based on the shots.~\cite{wang2022self} \\
 & Generate the missing answer based on the shots.~\cite{wang2022self} \\
 & Generate question-answer pairs based on the shots.~\cite{wang2022self} \\ \hline
Selection & Select high-quality data based on gradient information. ~\cite{xia2024less} \\ \hline
\end{tabular}}
\end{table}

\textbf{LLM-based Data Processing Strategy Generation.}
As shown in the left part of Figure~\ref{fig1}, LLM-AutoDP leverages  pre-trained LLMs as  agents to automatically generate DP strategies. 
We design a set of prompt templates and employ prompt engineering techniques to inject domain-specific knowledge into the LLMs' generation process. 
Guided by the prompts, LLMs produce diverse DP strategies that specify both the selection and sequencing of data processing techniques. 
These strategies are then executed in the model training environment, where they are iteratively refined based on performance feedback. 
A key challenge of this module is how the agent can effectively utilize  feedback to improve strategy generation. To address this, we propose a novel feedback mechanism based on group relative comparison, integrated with in-context learning, to guide LLMs toward  effective strategy formulation.

\textbf{Strategy Evaluation via Model Training and Evaluation.}
As shown in the right part of Figure~\ref{fig1}, to evaluate the sampled DP strategies, LLM-AutoDP applies the generated strategies to process the give data, followed by finetuning and evaluating a pre-trained model on the processed data. 
The evaluation results are then fed back to the strategy generation module to inform the next iteration of strategy optimization. 
This closed-loop interaction enables the DP strategies to dynamically adapt to the model training dynamics, thereby improving alignment between data processing and learning objectives. 
The central challenge of the evaluation module is how to achieve fast yet reliable strategy evaluation. 
To this end, in section ~\ref{sec:acce}, we introduce three key techniques to accelerate the evaluation process while preserving its fidelity.

\subsection{LLM-based Strategy Generation}

LLM-AutoDP leverages pre-trained LLMs as agents to automatically generate DP strategies. 
To achieve this, we employ prompt engineering techniques and carefully design a series of instruction templates. 
By incorporating task descriptions, input-output formats, and domain-specific  knowledge through prompts, we constrain the LLMs' generation process within a meaningful and practically feasible search space.
The strategy generation process runs iteratively.
In the initial round, strategies are generated based solely on our predefined instructions. 
In subsequent rounds, feedback from the strategy evaluation module is incorporated into the prompt to refine and optimize the generated strategies.

\textbf{Search Space.} { Considering four data processing teams, we explored the various permutation-combination scenarios when selecting 1, 2, 3, or 4 teams while taking order into account. Specifically, when selecting only one team, there are 4 options; when selecting two teams, there are 12 possibilities; when selecting three teams, there are 24 permutations; and when selecting all four teams, there are also 24 permutations. In total, this amounts to 64 different selection spaces. In addition, we have added an extra option: "No Processing Required for Original Data."}

\textbf{Initial Round.}
In the initial prompt, we first specify the objective: to design optimal DP strategies that improve data quality to achieve best model performance. We then describe the set of available DP operators and their corresponding functional groups, along with detailed explanations of each operator’s purpose and expected behavior. See Table~\ref{tab:ops} for more details on these operators.
Next, we instruct the LLMs to initialize multiple distinct DP strategies. Each DP strategy must clearly define the sequence of DP operators, resulting in a diverse set of initial DP strategies. We encourage the LLM to start with relatively small team compositions across different strategies to facilitate early exploration and understand the impact of different data processing methods and their orderings. This allows us to gather preliminary insights into the effectiveness of individual components and their interactions, forming a foundation for further optimization. For a detailed initial prompt, please refer to section ~\ref{prompt}.

\textbf{Iterative Optimization Rounds.}
Starting from the second round, the prompts provided to the LLMs not only include task instructions and operator descriptions but also incorporate performance feedback from previously evaluated strategies.
Let $\mathcal{S}^{(t)} = \{f_k^{(t)} \mid k = 1, \ldots, K^{(t)}\}$ denote the set of $K^{(t)}$ DP strategies generated in the $t$-th round. Let $R(\cdot)$ represent the strategy evaluation function. 
Then, the evaluation results for all strategies in $\mathcal{S}^{(t)}$ are $
\mathcal{R}^{(t)} = \{ r_k^{(t)} = R(f_k^{(t)}) \mid k = 1, \ldots, K^{(t)} \}.$
Additionally, we compute the baseline performance $r_0 = R(\text{None})$, which corresponds to training the model on the unprocessed dataset.
In the $(t+1)$-th round, the agent receives both $\mathcal{R}^{(t)}$ and $r_0$, and computes a normalized relative improvement score for each strategy as:
\begin{align}
    s_k^{(t)} &=  r_k^{(t)} - r_0, \; \forall 1 \leq k \leq K^{(t)}.
\end{align}

This score quantifies how much a given DP strategy improves or degrades model performance compared to the baseline. A positive score ($s_k^{(t)} > 0$) indicates a beneficial effect, while a negative score ($s_k^{(t)} < 0$)  suggests degradation. The higher the score, the more effective the strategy is at enhancing data quality for downstream training. We refine the prompt by injecting different DP strategies and their corresponding feedback scores as a group into the next round of prompting. Using inter-group relative comparison, we require the agent to compare and analyze the different DP strategy configurations and their feedback scores within the group. Through this process, the agent can identify superior data processing strategy configurations and, based on the insights gained, adjust the DP strategies, execution order, and the number of strategies in the next round.
The goal is to progressively refine the DP strategies toward optimal performance. 
When generating new strategies, the agent not only considers the absolute performance of previous strategies but also learns from their relative improvements.
If the agent determines that no further significant improvements can be made by adjusting group composition or ordering, it may terminate the iteration and return the most effective strategy found so far. An example of a prompt for the iterative optimization round can be found in section ~\ref{prompt}.

\subsection{Efficient Strategy Evaluation}
\label{sec:acce}

The strategy evaluation module aims to evaluate the effectiveness of the generated strategies and feed back the results to the agent for iterative optimization of the strategy.
To evaluate the performance of a candidate data processing strategy $f$, we insert it into the finetuning process of a pre-trained LLM $W_{pre}$, conduct finetuning for $N_{ft}$ epochs on the training set $\mathcal{D}$ that needs to be precessed, and then evaluate model performance on the validation set $\mathcal{D}_{val}$ using some evaluation function $\operatorname{E}_{val}$ to get results as feedback to agent.
This process can be formulated as follows:
\begin{align}
    R(f) = \operatorname{Eval}\bigg( \operatorname{FT}\big( W_{pre}, f(\mathcal{D}), N_{ft} \big), \mathcal{D}_{val}    \bigg)
\end{align}

However, since the evaluation process includes LLM training, it is very time-consuming, especially when the model has a large number of parameters. In addition, since we need to evaluate the strategy in each round of iterative optimization, which involves fine-tuning and testing of the LLM in each round, it is easy for the entire process to become unacceptably time-consuming. Therefore, the key problem of the evaluation module is how to achieve fast yet reliable strategy evaluation.
To solve this problem, we propose three technologies: Processing Target Selection, Distribution-Preserving Sampling, and Cache-and-Reuse Mechanism to reduce the time consumption of the strategy evaluation process.

\subsubsection{Training of the Binary Screening Model}\label{F-train}
A general approach to evaluating the quality of conversational data is to use prompt engineering to elicit answers from SOTA LLMs. However, due to their typically large parameter sizes, these models often have long inference times, which could introduce additional time overhead when applied within our framework. Inspired by knowledge distillation~\cite{xu2024survey}, we collected approximately 2 million pieces of medical and general conversational open-source data, 
which do not overlap with our training data. Following the prior work~\cite{chen2023alpagasus}, we scored the data on a scale of 1 to 5. To ensure diversity in the training data, we randomly sampled a certain amount of data from each scoring interval, forming a total of 200,000 training samples. Next, we used Qwen3-32B~\cite{yang2025qwen3} to annotate these 200,000 samples, assigning one of two labels: "High-quality data, no further processing required" or "Low-quality data, further processing needed." Finally, we trained Qwen2.5-7B~\cite{yang2025qwen3} on this dataset, minimizing the following loss function to obtain the final binary classification model for assessing data quality. This process aims to provide an efficient mechanism for accurately screening the quality of conversational data within a relatively short time frame.
\begin{align}
\mathcal{L}(\theta) = -\mathbb{E}_{(x,y) \sim \mathcal{D}_{clo}} \left[ \sum_{t=1}^{T} \log P_\theta(y_{t} \mid x, y_{<t}) \right]
\end{align}
Where $T$ represents the length of the target output sequence, and $D_{clo}$ denotes the 200,000 training samples we collected.

\subsubsection{Distribution-Preserving Sampling (DPS)} 
We denote the raw and unprocessed dataset as $\mathcal{D}_{{ori}}$. 
The key question of our evaluation is whether it is necessary to use the entire original dataset $\mathcal{D}_{{ori}}$ for model training. The  goal of model training and evaluation is to assess the performance of a given DP strategy and compare it against other candidates, to identify the most effective strategy. 
Therefore, rather than requiring the full dataset, we only need a subset of samples that sufficiently approximates the underlying data distribution.
To this end, we propose a sampling method that preserves the statistical properties of the original data distribution. 
Given a target sample size $N$ and a source dataset $\tilde{\mathcal{D}}$, we construct a reduced dataset $\mathcal{D}_s$ through iterative sampling until it contains exactly $N$ samples.
Initially, $\mathcal{D}_s$ is empty. At each iteration, we select one sample from $\tilde{\mathcal{D}} \setminus \mathcal{D}_s$ and add it to $\mathcal{D}_s$. To ensure the sampled subset retains the characteristics of the original data distribution, we employ an embedding model $E$~\cite{bge_embedding} to map each sample $x \in \tilde{\mathcal{D}}$ to a high-dimensional embedding vector $\mathbf{e}_x = E(x)$. We then select the sample whose embedding maximizes the similarity with all unselected samples, defined as:
\begin{align}\label{eq:sample}
    \tilde{\mathcal{D}}_{s}^{(i)} &= \tilde{\mathcal{D}}_{s}^{(i-1)} \cup \{x_i\} , \notag\\
    x_i &= \arg\max_{x}  \sum_{p \in \tilde{\mathcal{D}} \setminus \tilde{\mathcal{D}}_{s}^{(i-1)}}
    \frac{\mathbf{e}_{x}^\top {\mathbf{e}_{p}}}{\|\mathbf{e}_{x}\| \cdot \|{\mathbf{e}_{p}}\|}
\end{align}
We use the binary screening model $F$ to classify 
$\tilde{\mathcal{D}}$ into 
$\mathcal{D}_{{clean}}$ or $\mathcal{D}_{{noisy}}$. This process can be formulated as follows:
\begin{align}
    \mathcal{D}_{clean} &=\big\{ x | F(x)=0,\forall x \in  \tilde{\mathcal{D}} \big\},  \notag\\ \mathcal{D}_{noisy} &= \big\{ x | F(x)=1,\forall x \in  \tilde{\mathcal{D}}\}
\end{align}
To ensure that the sampled dataset accurately reflects the proportion of noisy samples of original data, we perform sampling separately on $\mathcal{D}_{{clean}}$ and $\mathcal{D}_{{noisy}}$ using Eq.~\ref{eq:sample}. Let $\tilde{\mathcal{D}}_{s,\text{clean}}$ and $\tilde{\mathcal{D}}_{s,\text{noisy}}$ denote the subsets sampled from $\mathcal{D}_{{clean}}$ and $\mathcal{D}_{{noisy}}$, respectively. 
The final sampled dataset $\mathcal{D}_s$ is then constructed as $\mathcal{D}_s = \tilde{\mathcal{D}}_{s,\text{noisy}} \cup \tilde{\mathcal{D}}_{s,\text{clean}}$.
This mechanism enables the preservation of the overall structure of the data manifold on both $\mathcal{D}_{clean}$ and $\mathcal{D}_{noisy}$. 
As a result, $\mathcal{D}_s$ provides a compact yet representative approximation of ${\mathcal{D}_{ori}}$, enabling efficient and reliable DP strategy evaluation.

\subsubsection{Processing Target Selection (PTS)}
When the agent generates a strategy, such as "Data Optimization --> Data Selection," we need to utilize an SOTA LLM to optimize all the data, and then compute gradient information, which inevitably leads to longer processing times. We consider that when sequentially processing the data according to the DP strategy, not all the data needs to be processed. Usually, there are some clean and high-quality data, which can be directly used for model training without additional processing. Therefore, we consider finding this part of the data in advance and skipping the processing on this data, thereby saving the time spent on processing this part of the data.
To this end, we use the binary screening model $F$, { which is trained using pipeline in section~\ref{F-train}},  to determine whether certain data in the current dataset needs to be optimized. Using this model, we can divide the dataset into two disjoint datasets, which contain clean data and data that need to be cleaned and optimized, respectively, as follows:
\begin{align}
    \mathcal{D}_{cur, clean} &=\big\{ x | F(x)=0,\forall x \in  \mathcal{D}_{cur} \big\},  \notag\\ \mathcal{D}_{cur, noisy} &= \big\{ x | F(x)=1,\forall x \in  \mathcal{D}_{cur}\}
\end{align}
where ${D}_{cur}$ represents the data processed in the current step according to the DP strategy.
When using a DP strategy $f$ for processing, we only process the data in $\mathcal{D}_{cur, noisy}$ or its subset one by one. The data in $\mathcal{D}_{cur, clean}$ can be directly used for model training.

\subsubsection{Cache-and-Reuse Mechanism (CRM)}
LLM-AutoDP employs an iterative optimization framework to discover effective data processing strategies. During this process, we observe that the same or partially overlapping strategies may be sampled across different rounds of optimization. This motivates us to design a cache-and-reuse mechanism to reduce redundant computation and accelerate strategy evaluation.

Specifically, in the $t$-th optimization round, the agent generate $K^{(t)}$ DP strategies as $\mathcal{S}^{(t)} = \{f_k^{(t)} \mid k = 1, \ldots, K^{(t)}\}$. 
For each strategy $f_k^{(t)} \in \{f_k^{(t)} \mid k = 1, \ldots, K^{(t)}\}$, we apply it to the sampled dataset $\mathcal{D}_s$ to obtain the processed dataset $\mathcal{D}_k^{(t)} = f_k^{(t)}(\mathcal{D}_s)$, which is stored in the dataset pool $\mathcal{C}^{(t)} = \{\mathcal{D}_k^{(t)} \mid k = 1, \ldots, K^{(t)}\}$.
In the $(t+1)$-th round, when evaluating a new strategy $f_k^{(t+1)}$, we first search the previously recorded strategy pool $\{f_j^{(i)} \mid 1 \le i \le t, 1 \le j \le K^{(i)}\}$ for the longest prefix of $f_k^{(t+1)}$. If such a prefix strategy $f_j^{(i)}$ exists and is identical to the prefix of $f_k^{(t+1)}$, we can decompose $f_k^{(t+1)}$ as:
\begin{align}
    f_k^{(t+1)} = f_{k,\text{prefix}}^{(t+1)} + f_{k,\text{suffix}}^{(t+1)} = f_j^{(i)} + f_{k,\text{suffix}}^{(t+1)},
\end{align}
where $f_{k,\text{prefix}}^{(t+1)}$ corresponds to the existing strategy $f_j^{(i)}$, and $f_{k,\text{suffix}}^{(t+1)}$ denotes the newly added operations.
Given this decomposition, the processed dataset $f_k^{(t+1)}(\mathcal{D}_s)$ is equivalent to applying the suffix $f_{k,\text{suffix}}^{(t+1)}$ to the precomputed dataset $\mathcal{D}_j^{(i)}$ as:
\begin{align}
    f_k^{(t+1)}(\mathcal{D}_s) = f_{k,\text{suffix}}^{(t+1)}(\mathcal{D}_j^{(i)}).
\end{align}
Therefore, instead of reprocessing the entire dataset from scratch using $f_k^{(t+1)}$, we can directly reuse $\mathcal{D}_j^{(i)}$ and only apply the suffix operations. 
This cache-and-reuse mechanism enables LLM-AutoDP to efficiently explore complex DP strategies while minimizing repeated computation, leading to lower overall training cost.

\section{Experiments}

\subsection{Experimental Settings}

\begin{table}[t]
\caption{
     This table demonstrates the optimal data processing strategy output by the agent after several rounds of iteration. In most experiments, LLM-AutoDP autonomously terminates after only 4 or 5 rounds. 
    }
    \label{tab:termail}
    \centering
    \scalebox{0.85}{\begin{tabular}{>{\raggedright\arraybackslash}p{2.3cm}c|cc}
        \toprule
        \multirow{2}{*}{Datasets} & \multicolumn{2}{c}{Termination Round} \\
        \cline{2-3}
        & Qwen3-32B & DeepSeek-R1-Distill-Llama-70B\\
        \midrule
        CMD &4 &4 \\
        \midrule
        cMedQA2 &5 &4 \\
        \midrule
        Medical-O1 &5 &5 \\
        \midrule
        Huatuo &4 &4 \\   
        \midrule
        Huatuo-100 &4 &3 \\
        \bottomrule
    \end{tabular}}
\end{table}

\subsubsection{Datasets, Metrics, and Models.}

We conduct experiments on medical QA datasets, as healthcare-related data is typically large in scale, highly sensitive, and often noisy, thereby presenting a pressing need for automatic data processing. 
Specifically, we use four medical dialogue and QA datasets: \texttt{cMedQA2}~\cite{8548603}, \texttt{Chinese-medical-dialogue} (CMD)~\cite{chinese_medical_data}, \texttt{Huatuo-26M-Lite}(Huatuo)~\cite{li2023huatuo26m}, and \texttt{Medical-O1-Reasoning-SFT}(Medical-O1)~\cite{chen2024huatuogpto1medicalcomplexreasoning}. 
Although these data have been simply cleaned during the collection process, our experiments show that it is still necessary to comprehensively process these data before using them. 
For each dataset, we randomly sample 20,000 instances for training and 1,000 for evaluation. 
In addition, we randomly sampled 100 data points from Huatuo-26M-Lite to form a small dataset, Huatuo-26M-Lite-100(Huatuo-100), to verify the effect of our method on a small dataset.
On the training sets, we apply various automatic data processing methods, and subsequently fine-tune three SOTA pre-trained LLMs on the processed data: \texttt{Qwen2.5-7B-Instruct}~\cite{yang2025qwen3}, \texttt{Llama3.1-8B-Instruct}~\cite{grattafiori2024llama}, and \texttt{Gemma-2-9B-Chat}~\cite{team2024gemma}. The fine-tuned models are then used to generate answers to the questions in the test sets. We follow the evaluation methods of medical LLMs as described in ~\cite{zhang2023huatuogpt, saab2024capabilities,yang2024zhongjing}.
We report \emph{win/tie/loss rates} of LLM-AutoDP against different baseline methods across all test sets. 
{ In the  results, "Our Wins", "Ties", and "Our Losses" represent the proportion of samples where our method performs better, the same, or worse than the baseline, respectively. 
The prompt for the win/tie/loss rates  follows prior work ~\cite{yang2024zhongjing} and details are shown in section ~\ref{prompt}}. Specifically, we pair models fine-tuned on data processed by LLM-AutoDP with those trained on data processed by other methods, and use two judge models \texttt{GPT-4}~\cite{achiam2023gpt} and \texttt{Baichuan-M1-14B-Instruct}~\cite{wang2025baichuan} to determine which model performs better on each question.

\subsubsection{Baselines.}
We compare LLM-AutoDP with several approaches for automated DP: 
\begin{itemize}
    \item[(1)] No-process. Using original unprocessed data. 
    \item[(2)] All-process. Using all data processing operators. We use all operators in Table~\ref{tab:ops} to form a strategy in the order of \textit{data $\text{data cleaning} \rightarrow \text{data optimization} \rightarrow \text{data generation} \rightarrow \text{data selection}$}. 
    \item[(3)] RS. Random search. Randomly selecting 5 strategies from all possible strategy combinations for evaluation and using the best one as the RS result. 
    \item[(4)] SELA~\cite{chi2024sela}, an AutoML framework that integrates LLMs with Monte Carlo Tree Search to automate traditional machine learning workflows. We modify the search target of SELA and only perform strategy search for the data processing process.
\end{itemize}

\begin{figure}[t]
    \centering
    \includegraphics[width=.70\linewidth]{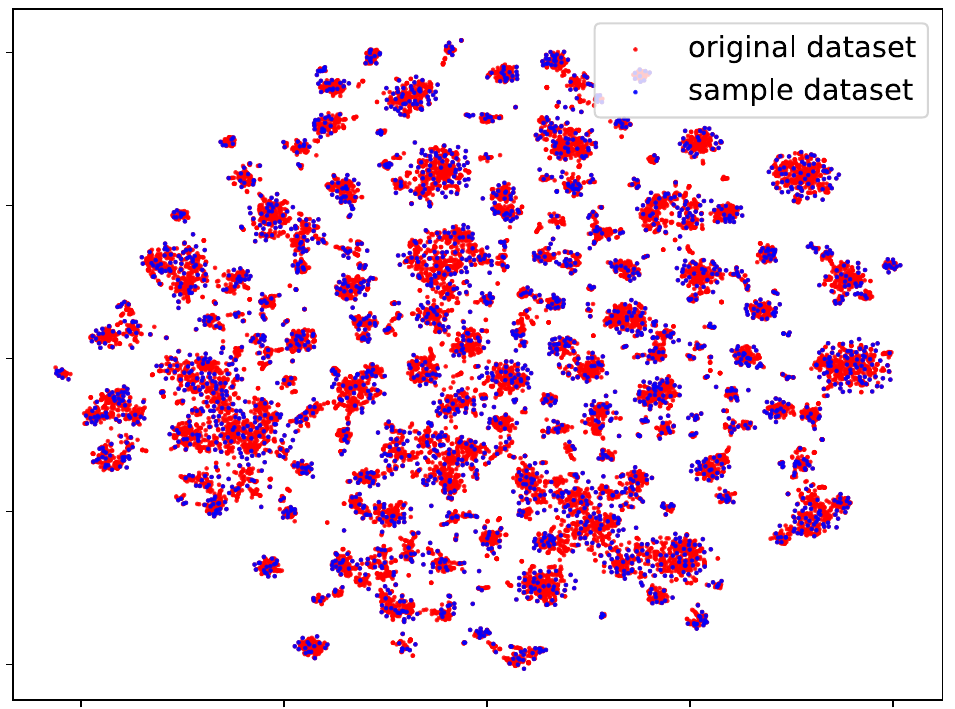}    \caption{Visualization of the distribution of the original data and the data sampled by 20\% on Huatuo‑26M‑Lite dataset. It can be seen that the distribution of the sampled data is consistent with that of the original data.}
    \label{fig:tsne}
\end{figure}

\subsubsection{Implementation Details.}
We utilize two agent models for strategy generation: \texttt{Qwen3-32B}~\cite{yang2025qwen3} and \texttt{DeepSeek-R1-Distill-Llama-70B}~\cite{deepseekai2025deepseekr1incentivizingreasoningcapability} (Due to the potential privacy concerns associated with the data, we prefer high-performance LLMs that can be deployed locally, rather than relying on callable APIs). The sampling temperature is set to 0.6. We sample 4 strategies at the initial round.

{ For sampling rate in DPS, previous study~\cite{dq} have shown that using 20\% of the data is sufficient to ensure the performance of the trained model. Below 20\% will negatively affect performance.
Therefore, we ensure that 20\% of the data from each dataset is sampled for DPS. We visualize the distribution of the sampled data and the original data in Figure~\ref{fig:tsne}. It can be seen that the two sets of distributions are consistent, indicating that our sampling process does not reduce the generalization ability.}

The fine-tuning process is conducted on {Qwen2.5-1.5B-Instruct} model.
For fine-tuning, we set the learning rate to 1e-5 and the batch size to 64. Each dataset was trained for 3 epochs. The AdamW optimizer was used for fine-tuning. We employed Swift~\citep{zhao2024swiftascalablelightweightinfrastructure} as the training platform and vLLM~\citep{kwon2023efficient} for inference. 
For iteration-based methods, SELA and our proposed LLM-AutoDP, we restrict the number of policy selection iterations to \textbf{at most 5 rounds}. This limitation is imposed due to the high computational cost for fine-tuning LLMs.
Exceeding this number would result in prohibitive time and resource consumption, making the policy selection process impractical.
All experiments are conducted with 16 NVIDIA A100 GPUs with 80G memory.

\subsection{Main Results Comparison}

\begin{table*}[bht]
\caption{Win/tie/loss results of LLM-AutoDP against different automatic data processing baselines using GPT-4 as judge. 
{"Our Wins","Ties", and "Our Losses" represent the proportion of samples where
our method performs better, the same, or worse than the baselines,
respectively.}
Results against No-Process result on Medical-O1-Reasoning-SFT are ties (Tie = 1.0), which is due to the high quality of the original data, leading the agents to determine that no additional processing is
required.}
    \label{tab:comp-gpt4}
    \centering
    \scalebox{0.9}{\begin{tabular}{>{\raggedright\arraybackslash}p{1.5cm}c|ccc|ccc|ccccc}
        \toprule
        \multirow{2}{*}{Datasets} & \multirow{2}{*}{LLM-AutoDP vs.} & \multicolumn{3}{c|}{Qwen2.5-7B} & \multicolumn{3}{c|}{Llama3.1-8B} & \multicolumn{3}{c}{Gemma-2-9B} \\
        \cline{3-11}
        & & Our Wins & Ties & Our Losses & Our Wins & Ties & Our Losses & Our Wins & Ties & Our Losses \\
        \midrule
        \multirow{4}{1.7cm}{Chinese-medical-dialogue} & No-Process & 0.8937 & 0.0057 & 0.1005 & 0.8549 & 0.0000 & 0.1451 & 0.8718 & 0.0247 & 0.1035 \\
        & All-Process & 0.5015 & 0.0922 & 0.4063 & 0.5062 & 0.0917 & 0.4021 & 0.5543 & 0.1217 & 0.3240 \\
        & RS & 0.7089 & 0.0461 & 0.2450 & 0.6931 & 0.0395 & 0.2674 & 0.7228 & 0.0288 & 0.2484 \\
        & SELA & 0.6945 & 0.0404 & 0.2651 & 0.6587 & 0.0606 & 0.2807 & 0.7125 & 0.0404 & 0.2471 \\
        \midrule
        \multirow{4}{1.7cm}{cMedQA2} & No-Process & 0.8463 & 0.0130 & 0.1407 & 0.8380 & 0.0087 & 0.1533 & 0.8818 & 0.0087 & 0.1094 \\
        & All-Process & 0.6583 & 0.0477 & 0.2940 & 0.6137 & 0.0359 & 0.3504 & 0.7130 & 0.0719 & 0.2151 \\
        & RS & 0.6733 & 0.0302 & 0.2965 & 0.6777 & 0.0302 & 0.2921 & 0.7335 & 0.0559 & 0.2106 \\
        & SELA & 0.5953 & 0.0377 & 0.3670 & 0.6233 & 0.0422 & 0.3345 & 0.6337 & 0.0325 & 0.3338 \\
        \midrule
        \multirow{4}{1.7cm}{Medical-O1-Reasoning-SFT} & No-Process & 0.0 & 1.0 & 0.0 & 0.0 & 1.0 & 0.0 & 1.0 & 0.0 & 0.0 \\
        & All-Process & 0.4724 & 0.0944 & 0.4332 & 0.4538 & 0.0713 & 0.4749 & 0.4766 & 0.0919 & 0.4315 \\
        & RS & 0.4969 & 0.0925 & 0.4106 & 0.5111 & 0.0833 & 0.4056 & 0.5252 & 0.0749 & 0.3999 \\
        & SELA & 0.4748 & 0.0933 & 0.4319 & 0.4698 & 0.0969 & 0.4333 & 0.5339 & 0.0738 & 0.3923 \\
        \midrule
        \multirow{4}{1.7cm}{Huatuo-26M-Lite} & No-Process & 0.6125 & 0.0225 & 0.3650 & 0.6250 & 0.0050 & 0.3700 & 0.6638 & 0.0174 & 0.3188 \\
        & All-Process & 0.5325 & 0.0475 & 0.4175 & 0.5418 & 0.0325 & 0.4257 & 0.5724 & 0.0475 & 0.3801 \\
        & RS & 0.5150 & 0.0500 & 0.4350 & 0.5590 & 0.0500 & 0.3910 & 0.5743 & 0.0500 & 0.3757 \\
        & SELA & 0.5325 & 0.0475 & 0.4175 & 0.5418 & 0.0325 & 0.4257 & 0.5724 & 0.0475 & 0.3801 \\
        \midrule
        \multirow{4}{1.7cm}{Huatuo-26M-Lite-100} & No-Process & 0.8312 & 0.0250 & 0.1438 & 0.8325 & 0.0025 & 0.1650 & 0.8237 & 0.0000 & 0.1763 \\
        & All-Process & 0.5575 & 0.1150 & 0.3275 & 0.5750 & 0.0275 & 0.3975 & 0.5773 & 0.0225 & 0.4002 \\
        & RS & 0.5950 & 0.1100 & 0.2950 & 0.6000 & 0.0250 & 0.3725 & 0.6137 & 0.0275 & 0.3588 \\
        & SELA & 0.6125 & 0.0825 & 0.3025 & 0.5475 & 0.0275 & 0.4250 & 0.5766 & 0.0425 & 0.3809 \\
        \bottomrule
    \end{tabular}}
\end{table*}

In Table~\ref{tab:comp-gpt4} and Table~\ref{tab:comp-baichuan}, we report the win/tie/loss rates of our proposed method, {LLM-AutoDP}, compared to other approaches, using GPT-4 and Baichuan-M1-14B-Instruct as judges, respectively.
The results demonstrate that LLM-AutoDP significantly outperforms the unprocessed baseline {No-Process} across most datasets. For instance, when fine-tuning various models on the {Chinese-medical-dialogue}, {cMedQA2}, and {Huatuo-26M-Lite-100} datasets under GPT-4 evaluation, the model trained on data processed by LLM-AutoDP achieves a win rate exceeding 80\% over the one trained on raw data. Similar trends are observed with Baichuan-based judgments.
An exception occurs in the {Medical-O1-Reasoning-SFT} dataset comparison, where all comparisons against {No-Process} result in ties (Tie = 1.0). This is attributed to the high quality of the original data, which leads the LLM to determine that no additional processing is necessary, resulting in identical training inputs.
Compared to the iterative method SELA, LLM-AutoDP achieves a higher win rate on the majority of datasets and models. Only in a few cases—such as fine-tuning {Llama3.1-8B} on {Medical-O1-Reasoning-SFT}—does LLM-AutoDP slightly underperform.
Notably, despite both being iterative strategy selection methods, LLM-AutoDP converges faster than SELA. In most experiments, LLM-AutoDP autonomously terminates after only 4 or 5 rounds, experiment results reported in table ~\ref{tab:termail}. In contrast, SELA remains far from convergence after 5 iterations~\cite{chi2024sela} and performs only marginally better than random search.
This highlights a key advantage of using LLMs as decision agents over traditional AutoML optimization techniques: LLMs can rapidly converge to more effective strategies. This efficiency is particularly valuable in LLM fine-tuning scenarios, where each iteration incurs substantial time and computational cost. Taken together, these results confirm that LLM-AutoDP offers a faster and more effective automated data processing solution.

\begin{table*}[ht]
\caption{Win/tie/loss results of LLM-AutoDP against different automatic data processing baselines using Baichuan-M1-14B-Instruct as judge. 
{"Our Wins","Ties", and "Our Losses" represent the proportion of samples where
our method performs better, the same, or worse than the baselines,
respectively.}
Results against No-Process result on Medical-O1-Reasoning-SFT are ties (Tie = 1.0), which is due to the high quality of the original data, leading the agents to determine that no processing is
needed.}
    \label{tab:comp-baichuan}
    \centering
    \scalebox{0.9}{\begin{tabular}{>{\raggedright\arraybackslash}p{1.5cm}c|ccc|ccc|ccccc}
        \toprule
        \multirow{2}{*}{Datasets} & \multirow{2}{*}{LLM-AutoDP vs.} & \multicolumn{3}{c|}{Qwen2.5-7B} & \multicolumn{3}{c|}{Llama3.1-8B} & \multicolumn{3}{c}{Gemma-2-9B} \\
        \cline{3-11}
        & & Our Wins & Ties & Our Losses & Our Wins & Ties & Our Losses & Our Wins & Ties & Our Losses \\
        \midrule
        \multirow{4}{1.7cm}{Chinese-medical-dialogue} & No-Process & 0.8694 & 0.0288 & 0.1018 & 0.8794 & 0.0288 & 0.0918 & 0.8709 & 0.0158 & 0.1133 \\
        & All-Process & 0.4021 & 0.2060 & 0.3919 & 0.4125 & 0.2488 & 0.3387 & 0.3967 & 0.2562 & 0.3471 \\
        & RS & 0.6816 & 0.1153 & 0.2031 & 0.6516 & 0.1334 & 0.2150 & 0.6771 & 0.1013 & 0.2216 \\
        & SELA & 0.6441 & 0.1311 & 0.2248 & 0.6693 & 0.1346 & 0.1661 & 0.6661 & 0.1214 & 0.2125 \\
        \midrule
        \multirow{4}{1.7cm}{cMedQA2} & No-Process & 0.8165 & 0.0778 & 0.1057 & 0.8429 & 0.0452 & 0.1119 & 0.8965 & 0.0166 & 0.0869 \\
        & All-Process & 0.5955 & 0.1520 & 0.2525 & 0.6069 & 0.1430 & 0.2501 & 0.6327 & 0.1541 & 0.2132 \\
        & RS & 0.6003 & 0.1683 & 0.2313 & 0.6148 & 0.1593 & 0.2259 & 0.6137 & 0.1889 & 0.1974 \\
        & SELA & 0.4584 & 0.2437 & 0.2979 & 0.4345 & 0.2297 & 0.3358 & 0.4467 & 0.2488 & 0.3045 \\
        \midrule
        \multirow{4}{1.7cm}{Medical-O1-Reasoning-SFT} & No-Process & 0.0 & 1.0 & 0.0 & 0.0 & 1.0 & 0.0 & 0.0 & 1.0 & 0.0 \\
        & All-Process & 0.5127 & 0.0963 & 0.3910 & 0.4769 & 0.0873 & 0.4358 & 0.5066 & 0.1003 & 0.3931 \\
        & RS & 0.4688 & 0.1017 & 0.4295 & 0.4793 & 0.0811 & 0.4396 & 0.4842 & 0.1178 & 0.3980 \\
        & SELA & 0.4421 & 0.0889 & 0.4690 & 0.4330 & 0.0108 & 0.5562 & 0.4688 & 0.0889 & 0.4423 \\
        \midrule
        \multirow{4}{1.7cm}{Huatuo-26M-Lite} & No-Process & 0.5401 & 0.1287 & 0.3312 & 0.5637 & 0.0925 & 0.3437 & 0.5931 & 0.0842 & 0.3227 \\
        & All-Process & 0.3913 & 0.1963 & 0.4124 & 0.4269 & 0.1717 & 0.4014 & 0.4543 & 0.1763 & 0.3694 \\
        & RS & 0.4525 & 0.2025 & 0.345 & 0.4444 & 0.1833 & 0.3723 & 0.4897 & 0.1818 & 0.3285 \\
        & SELA & 0.3913 & 0.1963 & 0.4124 & 0.4269 & 0.1717 & 0.4014 & 0.4543 & 0.1763 & 0.3694 \\
        \midrule
        \multirow{4}{1.5cm}{Huatuo-26M-Lite-100} & No-Process & 0.6400 & 0.2537 & 0.10625 & 0.9137 & 0.0425 & 0.0437 & 0.9300 & 0.020 & 0.050 \\
        & All-Process & 0.5213 & 0.19 & 0.2888 & 0.5525 & 0.1238 & 0.3237 & 0.5732 & 0.1168 & 0.3100 \\
        & RS & 0.5038 & 0.2912 & 0.205 & 0.565 & 0.1937 & 0.2413 & 0.5820 & 0.2027 & 0.2153 \\
        & SELA & 0.505 & 0.2662 & 0.2288 & 0.5475 & 0.1275 & 0.325 & 0.5690 & 0.1800 & 0.2510 \\
        \bottomrule
    \end{tabular}}
\end{table*}

\begin{figure*}[!th]
    \centering
    \includegraphics[width=0.85\linewidth]{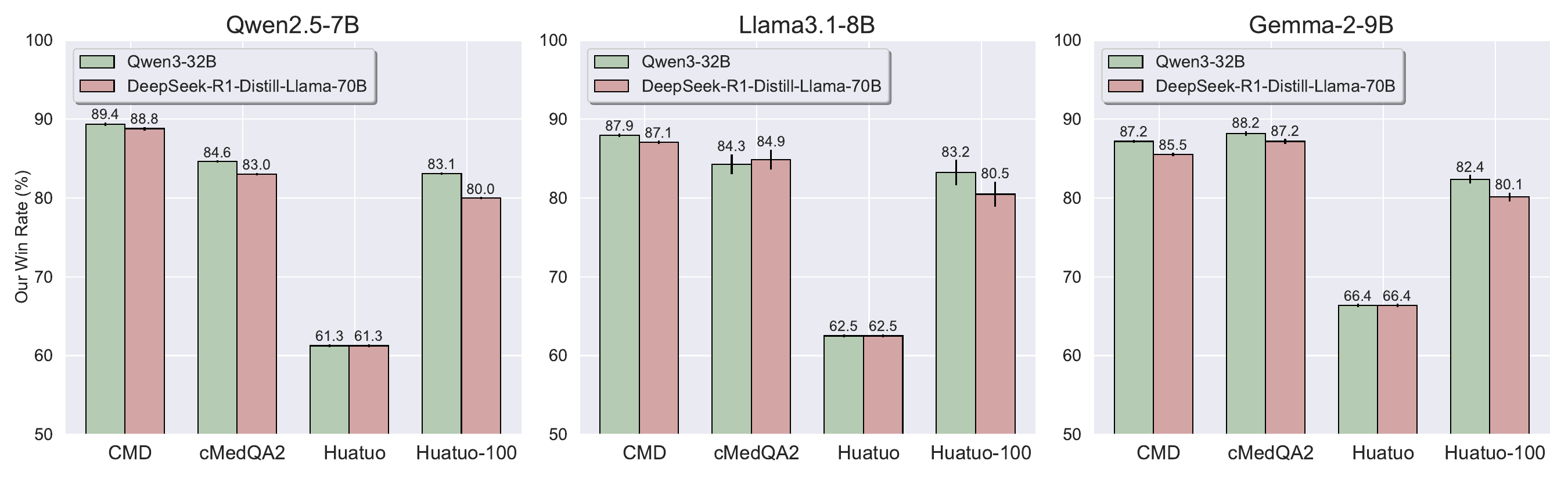}
    \includegraphics[width=0.85\linewidth]{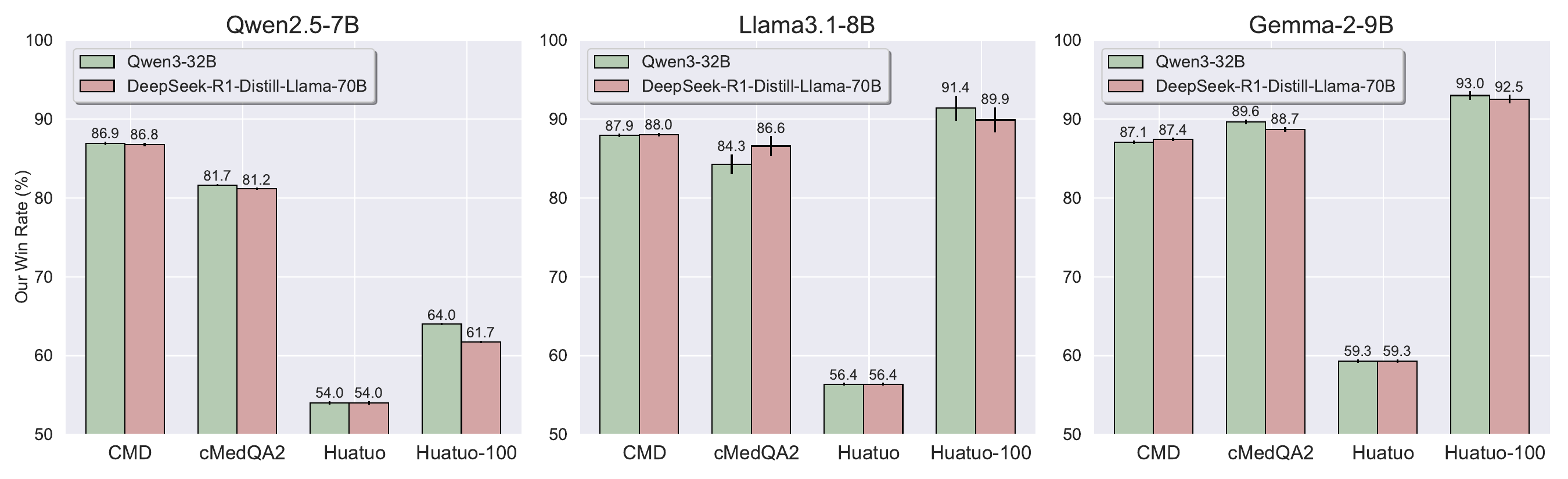}
    \caption{Results of using Qwen3-32B and DeepSeek-R1-Distill-Llama-70B as the agent models. 
    We repeat the experiments on each agent 5 times and record the win rate of the resulting fine-tuned models versus the models fine-tuned on the original, unprocessed data. 
    The colored columns and black bars indicate the average and standard deviation of repeated experiment results. 
    The top row and bottom row show results of using GPT-4 and Baichuan-M1-14B-Instruct as the judges, respectively.    
    }
    \label{fig:abl-agent}
\end{figure*}

\subsection{Efficiency Improvement of Acceleration Techniques}

To comprehensively evaluate the speedup effects of our proposed acceleration techniques (DPS, PTS, and CRM), we conduct ablation studies on four datasets. The results in Table~\ref{tab:efficient}, demonstrate the impact of different combinations of these techniques on the total time cost of the strategy search process.
As shown in Table~\ref{tab:efficient}, the baseline without any acceleration ({w/o DPS, PTS, CRM}) requires approximately 40–50 hours to complete the full search across all datasets.
This highlights the high computational burden associated with exhaustive exploration of the data processing strategy space. 
In contrast, introducing DPS alone significantly reduces the time required for strategy search by dynamically pruning unpromising branches early in the search process. 
For example, on the Chinese-medical-dialogue dataset, the total search time drops from 41.7 to 9.6 hours—a reduction of 76.9\%. Across all datasets, DPS consistently reduces the time cost by 76.3\%–78.7\%, demonstrating its effectiveness in efficiently narrowing down the dataset size for strategy searching.
When PTS is further introduced alongside DPS, we observe an additional reduction in search time by 2.2\%–3.3\%. Notably, on the Medical-O1-Reasoning-SFT dataset, the combination of DPS and PTS reduces the total search time from 50.0 to 4.3 hours, achieving a cumulative saving of 91.4\%. Similarly, on Huatuo-26M-Lite, the time is reduced to 6.7 hours, yielding an 81.9\% saving. 

The results indicate that PTS effectively reduces the time for data processing by filtering the samples that do not need processing.
Finally, integrating all three components (DPS, PTS, and CRM) achieves the highest speedup. On the Medical-O1-Reasoning-SFT dataset, the total search time is further reduced to just 2.5 hours—an impressive 95.0\% saving compared to the baseline. Across all four datasets, the combined use of the three techniques reduces the total search time by 88.2\%–95.0\%, clearly demonstrating their complementary roles in enhancing efficiency. 
In summary, these results validate the effectiveness of our proposed acceleration methods in substantially reducing the computational overhead of the strategy search process, without compromising the quality of discovered data processing pipelines.

\begin{table*}[ht]
\caption{Acceleration effects of the proposed efficient evaluation methods in reducing the time cost of the strategy search process. "{w/o DPS, PTS, CRM}" represents the baseline without any acceleration techniques applied. The column "Time (h)" indicates the total duration of the search process in hours, while "Save (\%)" reflects the percentage reduction in search time achieved by using the acceleration method. 
    }
    \label{tab:efficient}
    \centering
    \scalebox{0.9}{\begin{tabular}{l|cc|cc|cc|ccccccccccc}
        \toprule
        \multirow{2}{*}{Method} & \multicolumn{2}{c|}{Chinese-medical-dialogue} & \multicolumn{2}{c|}{cMedQA2} & \multicolumn{2}{c|}{Medical-O1-Reasoning-SFT} & \multicolumn{2}{c}{Huatuo-26M-Lite} \\
        \cline{2-9}
        & Time (h) & Save (\%) & Time (h) & Save (\%) & Time (h) & Save (\%) & Time (h) & Save (\%)  \\
        \midrule
        w/o DPS, PTS, CRM & 41.7 & 0 & 48.9 & 0. &50.0 & 0 & 37.2 & 0  \\
        + DPS & 9.6 & 76.9 & 10.4 & 78.7 & 11.2 & 77.6 & 8.8 & 76.3  \\
        + DPS, PTS & 8.7 & 79.1  & 9.4 &80.8  & 4.3 & 91.4 & 6.7 & 81.9  \\
        + DPS, PTS, CRM & 3.5 & 91.6 & 4.8 & 90.2  & 2.5 & 95.0 & 4.4  & 88.2 \\
        \bottomrule
    \end{tabular}}
\end{table*}

\subsection{Effects of Iterative Optimization}

To valid the effects of iterative optimization in {LLM-AutoDP}, we compare our LLM-AutoDP with a one-step strategy selection method that employs LLM for generation with greedy search for selection, denoted as LLM+greedy. In this approach, the LLM generates multiple candidate strategies at once, and the best-performing one is selected based on initial evaluation scores without any subsequent refinement or feedback-driven iteration.
Table~\ref{tab:comp-greedy} presents the win/tie/loss rates of LLM-AutoDP versus LLM+greedy across multiple datasets. The results show that LLM-AutoDP consistently outperforms the one-step method, achieving significantly higher win rates (ranging from 60\% to over 85\%) on most tasks. Tie rates are relatively low, indicating clear performance differentiation between the two approaches.
This comparison demonstrates the effectiveness of multi-round optimization in LLM-AutoDP. Unlike greedy search, which lacks iterative refinement, our method leverages feedback from previous rounds to progressively improve data processing strategies. This iterative mechanism enables the system to converge toward more effective pipelines, especially in complex and noisy data environments such as medical language modeling.
These findings underscore the importance of feedback loops in automated data processing for LLM fine-tuning, where a single static decision may fall short of capturing optimal solutions.

\begin{table*}[ht]
\caption{
    The effects of iterative optimization in LLM-AutoDP. 
    We compare LLM-AutoDP with a one-step strategy selection method (LLM+greedy) that employs LLM for strategy generation with greedy search for strategy selection. 
    We report the win/tie/loss rates of LLM-AutoDP versus LLM+greedy across multiple datasets and models. 
    }
    \label{tab:comp-greedy}
    \centering
    \scalebox{0.9}{\begin{tabular}{>{\raggedright\arraybackslash}p{3cm}c|ccc|ccc|ccccc}
        \toprule
        \multirow{2}{*}{Datasets} & \multirow{2}{*}{Judge} & \multicolumn{3}{c|}{Qwen2.5-7B} & \multicolumn{3}{c|}{Llama3.1-8B} & \multicolumn{3}{c}{Gemma-2-9B} \\
        \cline{3-11}
        & & Our Wins & Ties & Our Losses & Our Wins & Ties & Our Losses & Our Wins & Ties & Our Losses \\
        \midrule
        \multirow{2}{3cm}{Chinese-medical-dialogue} & GPT-4 
        & 0.7205& 0.0288& 0.2507	& 0.7011& 0.0317& 0.2672	& 0.7693& 0.0258& 0.2049 \\
        & Baichuan 
        & 0.6427& 0.1326& 0.2247	& 0.6442& 0.1095& 0.2463	& 0.6713& 0.0832& 0.2455 \\
        \midrule
        \multirow{2}{3cm}{cMedQA2} & GPT-4 
        & 0.7714& 0.0251& 0.2035	& 0.8266& 0.0025& 0.1709	& 0.8307& 0.0025& 0.1668 \\
        & Baichuan 
        & 0.7023& 0.1369& 0.1608	& 0.7815& 0.0979& 0.1206	& 0.7875& 0.0764& 0.1361 \\
        \midrule
        \multirow{2}{3cm}{Medical-O1-Reasoning-SFT} & GPT-4 
        & 0.4693& 0.1064& 0.4243	& 0.4612& 0.0915& 0.4473	& 0.4459& 0.0907& 0.4634 \\
        & Baichuan 
        & 0.4725& 0.0904& 0.4371	& 0.4642& 0.0842& 0.4516	& 0.4777& 0.1074& 0.4149 \\
        \midrule
        \multirow{2}{3cm}{Huatuo-26M-Lite} & GPT-4 
        & 0.575& 0.065& 0.36	& 0.6053& 0.0475& 0.3472	& 0.643& 0.0493& 0.3077 \\   
        & Baichuan 
        & 0.4538& 0.205& 0.3412	& 0.45& 0.2263& 0.3237	& 0.4927& 0.1967& 0.3106 \\
        \midrule
        \multirow{2}{3cm}{Huatuo-26M-Lite-100} & GPT-4 
        & 0.6825& 0.05& 0.2675	& 0.6075& 0.0075& 0.385	& 0.6415& 0.0125& 0.346 \\
        & Baichuan 
        & 0.6188& 0.1925& 0.1887	& 0.6813& 0.1237& 0.195	& 0.6717& 0.1461& 0.1822 \\
        \bottomrule
    \end{tabular}}
\end{table*}

\subsection{Effects of using different LLMs as agents }

In the above experiments, we employ \texttt{Qwen3-32B}~\cite{yang2025qwen3} as the agent model for strategy generation. 
To assess whether LLM-AutoDP's performance is sensitive to the choice of LLM agent, we replace Qwen3-32B with \texttt{DeepSeek-R1-Distill-Llama-70B}~\cite{deepseekai2025deepseekr1incentivizingreasoningcapability}.
We repeat the experiments on both models 5 times and record the win rate of the resulting fine-tuned models when strategies were generated using each agent, relative to models trained on the original, unprocessed data.
We present the average and standard deviation of repeated experiment results in Figure~\ref{fig:abl-agent}.
The results in Figure~\ref{fig:abl-agent} demonstrate that both LLMs are effective agents for generating high-quality DP strategies across various datasets and base models. 
While \texttt{Qwen3-32B} achieves slightly better performance than \texttt{DeepSeek-R1-Distill-Llama-70B}, especially on the \texttt{Huatuo-100} dataset, the average performance difference across other datasets is negligible. 
This indicates that LLM-AutoDP is relatively robust to the choice of agent model, suggesting a low dependency on specific LLM architectures.
Furthermore, we observe that the standard deviation across multiple experimental runs is consistently small.
This suggests that LLM-AutoDP also exhibits robustness to the sampling process of the agent LLMs' strategy generation, demonstrating stability in its automated pipeline.

\begin{figure*}[!th]
    \centering
    \includegraphics[width=0.85\linewidth]{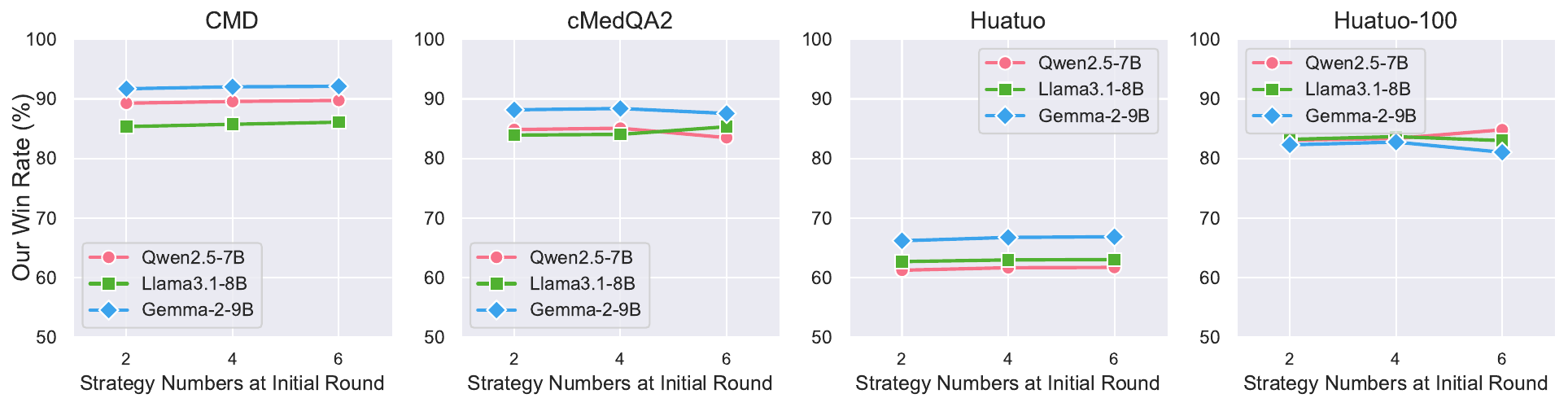}
    \includegraphics[width=0.85\linewidth]{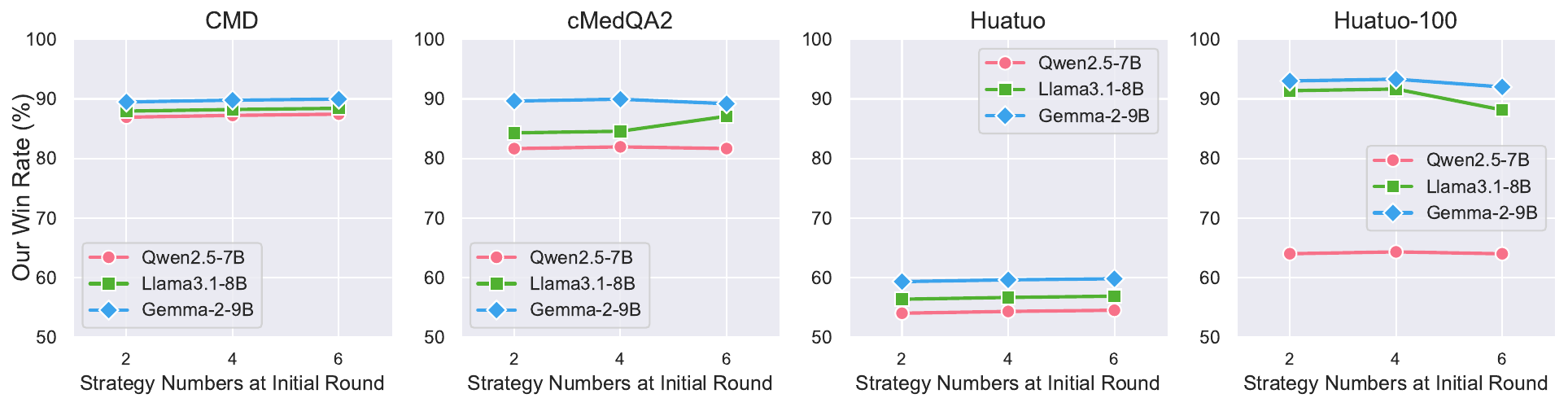}
    \caption{
    Results of using different numbers of strategies in the initial round.
    We use Qwen3-32B as the agent model and report the win rate of the resulting fine-tuned models versus the models fine-tuned on the original, unprocessed data. 
    The top row and bottom row show results of using GPT-4 and Baichuan-M1-14B-Instruct as the judges, respectively. 
    }
    \label{fig:abl-init}
\end{figure*}

\subsection{Impact of initial strategy numbers}

In the first round of LLM-AutoDP, we inform the agents of the initial strategy numbers via the prompt. 
This parameter directly influences the exploration space during the initial phase. 
To investigate whether this parameter affects the final strategy quality, we conducted experiments with different initial strategy sampling counts $\{2, 4, 6\}$.
Figure~\ref{fig:abl-init} demonstrate the performance across various fine-tuned models and datasets. 
We find that varying the initial sampling count has minimal impact on the final strategy effectiveness. Specifically, all fine-tuned models show performance differences of no more than 0.5\% across the three sampling settings. Across four datasets, the performance fluctuations due to different initial sampling counts remain below 1\%. 
The key reason for this robustness lies in the adaptive nature of LLM-AutoDP during subsequent iterations.
Although the initial exploration space differs, we observe that the framework automatically adjusts both the number of exploration rounds and the number of strategies explored per round. 
For instance, with an initial sampling of 6 strategies, LLM-AutoDP converges within 3 iterations, whereas it requires 5 iterations when starting with only 2 strategies.
These observations highlight the robustness of the LLM-AutoDP framework to the initial exploration configuration, demonstrating its strong adaptability to diverse initial settings without compromising the quality of the resulting data processing strategies.

\section{Conclusion and Future Work}

We propose LLM-AutoDP to address the challenge of automating data processing for LLM fine-tuning without exposing raw, potentially sensitive training data. 
Our method apply LLMs as  agents to iteratively generate and refine DP strategies through alternating phases of strategy generation and evaluation.
To accelerate the computationally expensive evaluation phase, we introduce three key techniques.
These techniques significantly reduce time consumption while preserving the effectiveness of the discovered strategies.
Experiments across various datasets and models show that LLMs fine-tuned on data processed by our framework achieve over 80\% win rates compared to models trained on unprocessed data and outperform AutoML baselines based on LLM agents with approximately a 65\% win rate. 
Our proposed acceleration techniques reduce the total search time by up to $10\times$, demonstrating both high efficiency and strong performance.
As future work, we plan to extend LLM-AutoDP to support multi-modal data and explore its applicability in other high-stakes domains. 
Another future direction to combine  privacy-preserving mechanisms into the framework to further enhance data security.
{
For instance, differential privacy can be incorporated into the training process to prevent the model from leaking sensitive information. In such cases, the effectiveness of LLM-AutoDP warrants further investigation.
}

\section{appendix}
\subsection{Prompts used in LLM-AutoDP}\label{prompt}

\textbf{(1) Prompt for agent LLMs at initial round:}
\begin{shaded} 
    You are an intelligent assistant specializing in handling training data for large models. The current task is to achieve a specified level of accuracy by adjusting the quality of the data, but this must be done in "black-box mode" — meaning you cannot view the specific content of the data or know in advance what issues exist within it. To accomplish the goal, you can coordinate the following four professional teams to work collaboratively:

    (1) Data Cleaning Team: Responsible for basic data purification.
    
    (2) Data Generation Team: Responsible for supplementing new data.
    
    (3) Data Optimization Team: Responsible for improving data quality.
    
    (4) Data Selection Team: Responsible for filtering high-quality data.

    For more details on each team's specific functions, refer to the "Team Capability Overview."
    
    Although you cannot preview the problems within the data, you can combine the teams like assembling building blocks — deciding which teams to use and planning the sequence of their work (the order will affect the final outcome). Through continuous iteration, you need to identify the optimal combination of teams to process the training data.

    \#\#\# Team Capability Overview:
    
    (1) ** Data Cleaning Team **

    $\bullet$ Capabilities: Data purification, including deduplication, removal of special characters, filtering data with inappropriate length, noise reduction, and elimination of repeated words within individual data entries.
    
    (2) ** Data Generation Team **
    
    $\bullet$ Capabilities: Generating missing questions or answers, as well as creating new data.
    
    (3) ** Data Optimization Team **
    
    $\bullet$ Capabilities: Improving data quality by optimizing existing data, enhancing the accuracy of Q\&A pairs, improving text fluency, readability, conciseness, and ensuring compliance with content standards.
    
    (4) ** Data Selection Team **
    
    $\bullet$ Capabilities: Selecting high-quality data.

    \#\#\# Task Execution Instructions:
    
    First Round (Initialization): Initialize multiple different team combinations (no more than four)
    
    $\bullet$ Based on your experience, initialize multiple distinct combinations of teams.
    
    $\bullet$ Each combination can consist of a single team or multiple teams.
    
    $\bullet$ The sequence of the teams' work must be clearly defined for each combination.
    
    $\bullet$ During the first round of initialization, it is recommended to test combinations with fewer teams to determine the initial impact of different teams and their working order. This will help accumulate baseline data for subsequent iterative optimization.
    
    \#\#\# Example of Combinations:
    
    \#\#\#Combination[1]\#\#\#
    
    $\bullet$ Data Cleaning Team
    
    \#\#\#Combination[2]\#\#\#
    
    $\bullet$ Data Cleaning Team, Data Generation Team
    
    Output Format:
    
    Please strictly follow the format below to output the results:
    
    \#\#\#Combination[1]\#\#\#
    
    $\bullet$ List the team names in order of execution, separated by commas.
    
    ...
    
    \#\#\#Combination[n]\#\#\#
    
    $\bullet$ List the team names in order of execution, separated by commas.
    
    \#\#\#Reasons for Different Combinations\#\#\#
    
    $\bullet$ Explain why these combinations were chosen.

    Round 1:
    
    Now, proceed with the first step and initialize multiple different team combinations (no more than four).

\end{shaded}

\textbf{(2) Prompt for agent LLMs at iterative optimization rounds:}
\begin{shaded} 
    Round 2 (Iterative Optimization): Evaluate Feedback Scores of Combinations and Adjust Team Configurations and Work Order; Changes in the Number of Combinations Are Allowed

    $\bullet$ Obtain feedback scores for different combinations. A higher score for a combination indicates that the data processed by this combination is more suitable for large model training.

    $\bullet$ The feedback scores reflect the performance differences between the processed training data and the original data. Therefore, the score results may indicate either positive improvement (positive values) or performance degradation (negative values).

    $\bullet$ Adjust team configurations and work order based on the feedback scores from the previous round. The goal is to progressively optimize the data processing effect to achieve the best possible outcome.

    $\bullet$ You are allowed to change the number of combinations in this round. That is, the number of combinations in this round can differ from the previous round.

    $\bullet$ When adjusting team configurations, pay attention not only to the scores of different combinations but also to the relative changes in scores among different combinations.

    $\bullet$ If, during the iteration process, you determine that no further adjustments to team configurations or work order are needed, you can stop the iteration and provide the most suitable team combination. Mark this combination with the label 【Best Team】 to indicate it as the final choice.

    $\bullet$ If, during the iteration process, you attempt different combinations but the feedback scores are consistently close to zero, it indicates that the original data does not require any processing. In this case, output 【No Processing Required for Original Data】 and stop the iteration.

    $\bullet$ Duplicate teams are not allowed within the same combination, and the number of teams in a single combination must not exceed four.

    Repeat Round 2 until you determine that no further adjustments to team configurations or work order are necessary.

    The feedback scores for the different team combinations in Round {} are as follows:

    1.\#\#\#Combination[1]\#\#\#

    Feedback Score: {}

    2.\#\#\#Combination[2]\#\#\#

    Feedback Score: {}

    3.\#\#\#Combination[3]\#\#\#

    Feedback Score: {}

    4.\#\#\#Combination[4]\#\#\#

    Feedback Score: {}

    \#\#\# Round {}:

    Now, proceed with Step {}, evaluate the feedback scores of the combinations, and adjust the team configurations and work order. Changes in the number of combinations are allowed.

\end{shaded}

\textbf{(3) Prompt for win/wie/loss rate judge models:}
\begin{shaded} 
    As a professional medical evaluator, please evaluate the following two doctors' responses to the same medical question.
    
    Question:{}
    
    Response 1:{}

    Response 2:{}
    
    The evaluation criteria are prioritized in the following order: Accuracy of the doctor's response, Safety, Fluency, and Conciseness. The specific definitions are as follows:
    
    Evaluation Criteria:
    
    1. Accuracy of the Doctor's Response: The doctor should accurately understand the patient's question and provide a scientific and correct answer.
    
    2. Safety: The doctor must adhere to laws, regulations, ethics, and professional conduct when answering.
    
    3. Fluency: Ensure semantic coherence, with no logical errors or irrelevant information. Maintain a friendly and enthusiastic tone in the response.
    
    4. Conciseness: Clearly and concisely explain complex medical knowledge. Avoid overly redundant content in the dialogue.
    
    Note: The evaluation must be based on the importance ranking of Accuracy > Safety > Fluency > Conciseness. In case of conflicts, prioritize the higher-ranking criterion.
    
    You need to select your evaluation result from the following three options: [Response 1 wins compared to Response 2, Response 1 ties with Response 2, Response 1 loses compared to Response 2].
    
    Your output must strictly follow the format below:
    
    Evaluation Result:
    
    Only provide the selected evaluation result here.
\end{shaded}

\subsection{Experiments on Law Dataset}

\begin{table}[h]

\caption{
    The results of LLM-AutoDP on DISC-Law-SFT dataset.
    We report the win/tie/loss rates of LLM-AutoDP versus No-process baseline.
    }
    \label{tab:law}
    \centering
    \scalebox{0.9}{\begin{tabular}{lc|ccccccccccc}
        \toprule
        {Models} & {Judge} & Our Wins & Ties & Our Losses \\
        \midrule
        \multirow{2}{3cm}{Qwen2.5-7B} & GPT-4 
        & 0.9007& 0.0145& 0.0848 \\
        & LawLLM 
        & 0.8419& 0.0625& 0.0956	 \\
        \midrule
        \multirow{2}{3cm}{Llama3.1-8B} & GPT-4 
        & 0.8466& 0.000& 0.1534\\
        & LawLLM 
        &0.8664& 0.0394& 0.0942 \\
        \midrule
        \multirow{2}{3cm}{Gemma-2-9B} & GPT-4 
        & 0.8941& 0.0024& 0.1035\\
        & LawLLM 
        & 0.8327& 0.1049& 0.0624	 \\
        \bottomrule
    \end{tabular}}
\end{table}

In addition to medical data, we also validate the effectiveness of LLM-AutoDP on law dataset.
We evaluate LLM-AutoDP on the \texttt{DISC-Law-SFT} dataset~\cite{lawdata}. 
We compare the performance of models trained on data processed by LLM-AutoDP with those trained on the original data. The evaluation is conducted using GPT-4 and a domain-specialized legal model, \texttt{LawLLM-7B}~\cite{lawllm}, as judges.
As shown in Table~\ref{tab:law}, the models trained on LLM-AutoDP-processed data outperform those trained on the original data in approximately 90\% of the queries. 

\subsection{An Example of Strategy Optimizations}

\begin{figure}[!h]
    \centering
    \includegraphics[width=1\linewidth]{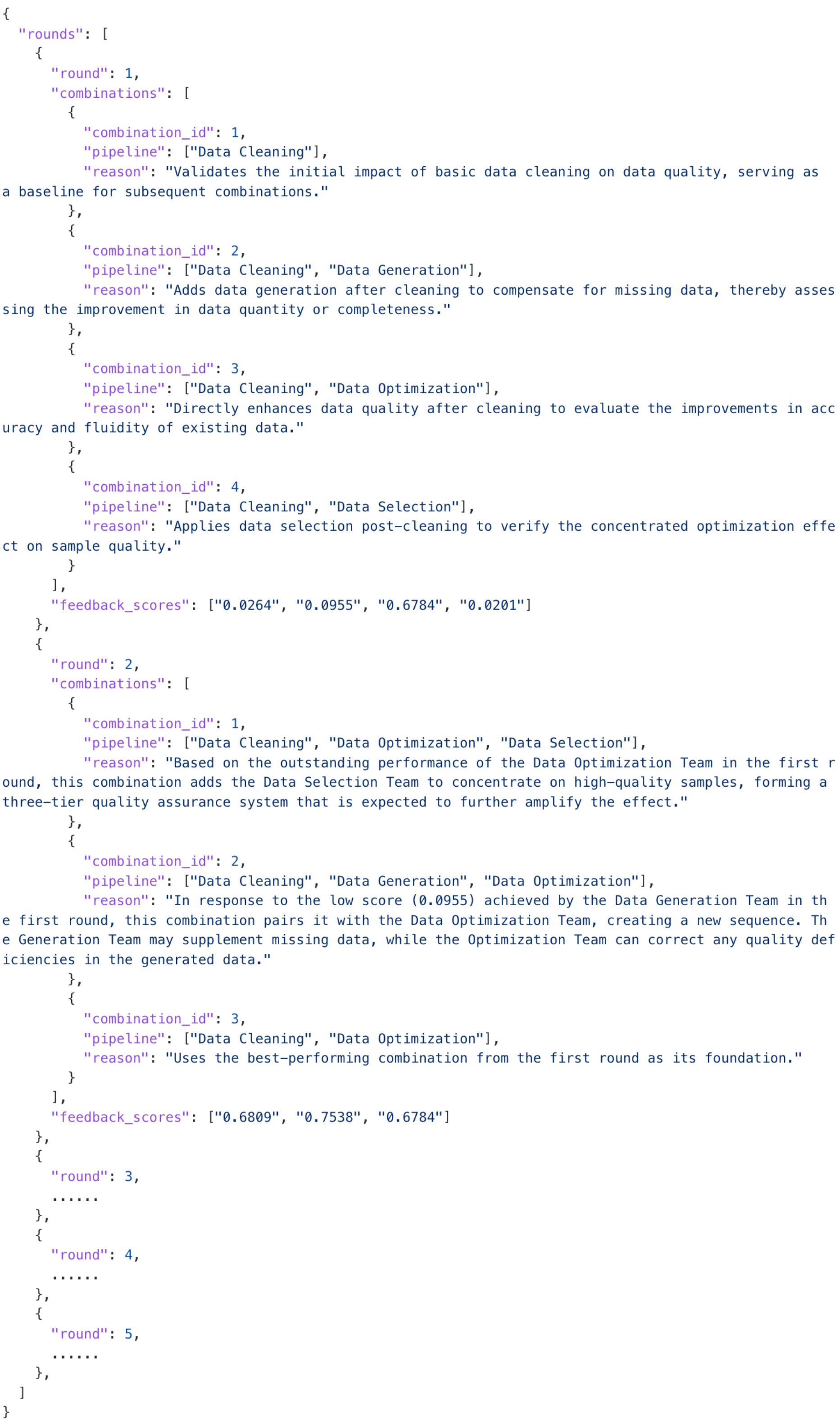}
    \caption{A step-by-step example of strategy optimization using LLM-AutoDP. A total of 5 rounds of iterations are performed, but only the details of 2 rounds are shown due to space imitation.}
    \label{fig:example}
\end{figure}

We present an example of the step-by-step optimization process by LLM-AutoDP on cMedQA2 dataset in Figure~\ref{fig:example}.

\color{black}

\clearpage

\bibliographystyle{ACM-Reference-Format}
\bibliography{sample}

\end{document}